\documentclass[journal]{IEEEtran}

\usepackage{graphicx}  
\usepackage{multirow}
\usepackage{times}
\usepackage{epsfig}
\usepackage{graphicx}
\usepackage{amsmath}
\usepackage{amssymb}
\usepackage{color}
\usepackage{array}
\usepackage{makecell}

\ifCLASSINFOpdf
\else
\fi
\begin{document}
%
\title{Learning to Compare Relation: Semantic Alignment for Few-Shot Learning}
%
%
%

\author{Congqi~Cao,~\IEEEmembership{Member,~IEEE,}
        Yanning~Zhang,~\IEEEmembership{Senior~Member,~IEEE}
\thanks{C. Cao and Yannning Zhang are with the National Engineering Laboratory for Integrated Aero-Space-Ground-Ocean Big Data Application Technology, School of Computer Science, Northwestern Polytechnical University, Xi'an 710129, China (e-mail: congqi.cao@nwpu.edu.cn; ynzhang@nwpu.edu.cn).}
}

\maketitle

\begin{abstract}
Few-shot learning is a fundamental and challenging problem since it requires recognizing novel categories from only a few examples. The objects for recognition have multiple variants and can locate anywhere in images. Directly comparing query images with example images can not handle content misalignment. The representation and metric for comparison are critical but challenging to learn due to the scarcity and wide variation of the samples in few-shot learning. In this paper, we present a novel semantic alignment model to compare relations, which is robust to content misalignment. We propose to add two key ingredients to existing few-shot learning frameworks for better feature and metric learning ability. First, we introduce a semantic alignment loss to align the relation statistics of the features from samples that belong to the same category. And second, local and global mutual information maximization is introduced, allowing for representations that contain locally-consistent and intra-class shared information across structural locations in an image. Furthermore, we introduce a principled approach to weigh multiple loss functions by considering the homoscedastic uncertainty of each stream. We conduct extensive experiments on several few-shot learning datasets. Experimental results show that the proposed method is capable of comparing relations with semantic alignment strategies, and achieves state-of-the-art performance.
\end{abstract}

\begin{IEEEkeywords}
few-shot learning, relation modeling, semantic alignment.
\end{IEEEkeywords}

%
\IEEEpeerreviewmaketitle

\section{Introduction}

In practical application scenarios, annotation is not easy to obtain. The demand for a large number of annotated samples restricts the application scope of deep learning algorithms. Few-shot learning has attracted increasing attention recently due to its potential wide applications in practice \cite{Matching_Networks_NIPS16, Prototypical_NIPS17, Compare_CVPR18, graph_matching_ECCV18, adaptive_metric_NIPS18, Task_Relevant_CVPR19, adaptive_posterior_ICLR19, LGM_ICML19, f_VAEGAN_CVPR19, Hallucination_CVPR19}.

However, it is challenging for machine learning systems to learn novel concepts from a few samples like human beings. Researchers try to solve it from different perspectives.
The existing approaches can be generally categorized into three classes: the approaches based on matching networks, the approaches based on meta-learner optimization, and the approaches based on data augmentation.
Matching network based approaches \cite{Matching_Networks_NIPS16, Prototypical_NIPS17, Compare_CVPR18, graph_matching_ECCV18, adaptive_metric_NIPS18, Local_Descriptor_CVPR19, Dense_Classification_CVPR19, Task_Relevant_CVPR19} rely on the idea that samples from the same category are more similar than samples from different categories. It is important to choose an appropriate feature space and metric criterion to measure the similarity.
Meta-learner optimization based approaches \cite{MANN_ICML16, bertinetto2016learning, Optimization_ICLR17, MAML_ICML17, TAML_CVPR19, adaptive_posterior_ICLR19, LGM_ICML19} treat few-shot learning as a fast learning and optimization problem. These models can be seen as composed of two parts, \textit{i.e.}, meta-learner and learner. They focus on optimizing the meta-learner which can determine the initialization parameters, structure or learning strategy of the learner.
Besides these approaches, data augmentation \cite{Model_Regression_ECCV16, f_VAEGAN_CVPR19, Hallucination_CVPR19, patch_sampler_CVPR19, meta_transfer_CVPR19, knowledge_transfer_CVPR19} is another way to solve few-shot learning problems. Increasing the number of training samples or transfer the knowledge learned from large-scale datasets can improve the few-shot recognition performance.
We follow the first direction, which has been widely used when the samples with labels are extremely limited.

For matching based approaches, metric measures based on image-level features are usually used.
Vinyals \textit{et al.} \cite{Matching_Networks_NIPS16} used a convolutional network followed by a Long-Short Term Memory (LSTM) as the context embedding function for images. A weighted nearest-neighbor classifier based on cosine distance was utilized for label prediction.
Snell \textit{et al.} \cite{Prototypical_NIPS17} represented each class by the mean of its support examples' convolutional embeddings and performed classification by computing Euclidean distances to the prototypes of each class.
Sung \textit{et al.} \cite{Compare_CVPR18} proposed to learn a deep distance metric over the concatenated feature maps of query and support examples with a network for classification. These methods achieved promising results on few-shot learning task.
However, the objects for recognition have multiple variants and can locate anywhere in images. Directly comparing two images using image-level features cannot handle content misalignment of the samples well.
To solve this issue, Li \textit{et al.} \cite{Local_Descriptor_CVPR19} used a local descriptor based measure via k-nearest neighbor search over the local descriptors of convolutional feature maps.
Hao \textit{et al.} \cite{SAML_ICCV19} calculated the distance of each local region pairs and used attention mechanism to put more weights on the semantically relevant pairs for comparison.
These local descriptor based methods are robust to translation. However, they are still constrained by static appearance feature comparison which is sensitive to absolute value, transformation, object variations and noises.

Besides appearance information, relation information is another crucial factor in semantic description which is robust to content misalignment. It describes the relationship among the local elements inside an image. The specific form can be a kind of transformation, correlation, consistency, etc. It has been successfully applied in fine-grained visual recognition, multi-modal learning, and action recognition tasks \cite{lin2015bilinear, ARTNet_CVPR18, SSA_CVPR19}.
However, it has not been explored well in few-shot learning area, which faces the difficulty of data scarcity and semantic misalignment in appearance.
Since directly coupling the information of appearance and relation together in one linear combination network adds the difficulty for modeling and increases the over-fitting risk \cite{ARTNet_CVPR18}, which has not been solved well even in the tasks with a large amount of training data, we propose to learn and compare appearance and relation features separately and complementarily for few-shot learning.

\begin{figure}[!t]
\centering
\includegraphics[width=\columnwidth]{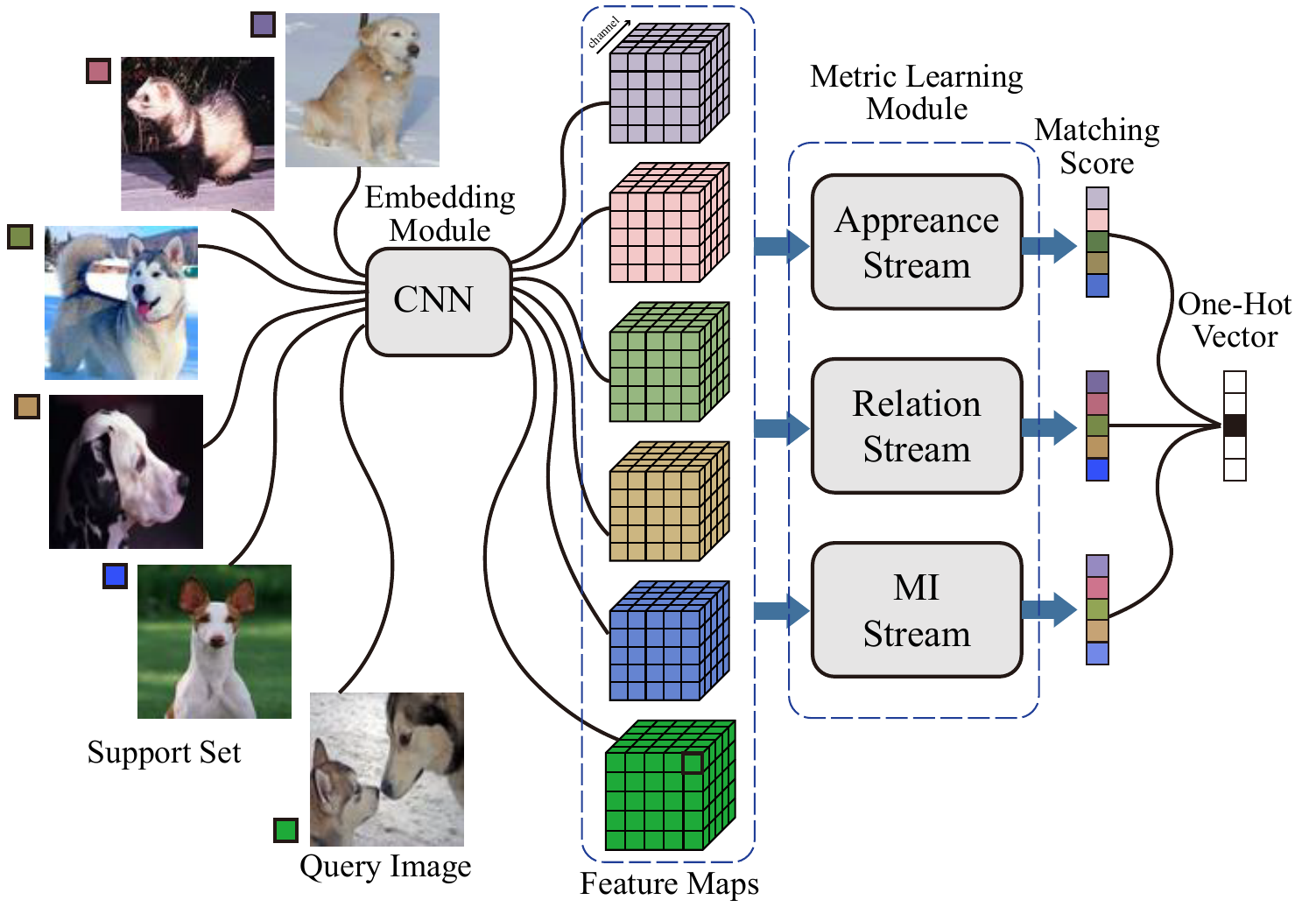}
\caption{Framework of our model. Images are firstly fed forward into an embedding module for feature extraction. Then a metric learning module compares the image representation pairs and generate a set of matching scores. At last, a class label prediction is obtained by aggregating the matching scores with a fusion module.}
\label{fig:framework}
\end{figure}

In this paper, we present a novel semantic alignment model for few-shot learning.
The framework of our proposed method is illustrated in Figure \ref{fig:framework}. It mainly consists of three modules: an embedding module, a metric learning module, and a fusion module. The metric learning module is composed of an appearance stream, a relation stream and a mutual information (MI) stream.
Among them, the relation stream and the MI stream are introduced for semantic alignment and relation comparing.
Given a set of labelled example images and a query image without label,
firstly, the images are fed forward into the embedding module, which is a convolutional neural network (CNN) in our paper, for feature extraction. Then the metric learning module compares the deep learned representations of the images to generate a set of matching scores.
Finally, through the fusion module, the predicted class label of the query image is obtained by aggregating the matching scores using homoscedastic uncertainty, which can be interpreted as stream-dependent weighting.
The propose method can acquire new examples rapidly while providing excellent generalisation from common examples.

Specifically, besides the appearance stream which directly compares the local appearance information between feature pairs,
inspired from the spatiotemporal semantic alignment loss \cite{SSA_CVPR19} introduced to align the feature content from different modalities,
we compare query image and example image pairs first by aligning the semantics of the deep representations. We do this by enforcing them to share a common correlation matrix across the deep features of all the samples belonging to the same category. This is done by minimizing the distance between their correlation matrices in the training stage. Since this mechanism can compare the relations of the elements inside an image, we call this a relation stream in our network.
In order to take both the positional information and the style information into account, we apply the semantic alignment loss to the correlation matrix across spatial positions as well as the correlation matrix across channels.

Furthermore, we propose to improve the network's representational capacity by using local-global mutual information, which can solve misalignment and compare relation from another perspective.
Maximizing mutual information between input and the learned representations has been widely used in unsupervised representation learning \cite{mutual_information_ICLR19, sun2019contrastive}.
For few-shot learning, we argue that
the mutual information between the images from the same class should be large, while the mutual information between the images from different classes should be small. Given a query image, even if its content is misaligned with that of the example image, a good learning algorithm should be able to conjure up the whole thing through seeing a part of it. Hence, we introduce a MI stream in our network to maximize the mutual information between local patches and the global representation of examples with the same class label. This encourages the embedding module to prefer information that is shared across the same class regardless of locality.

In order to combine the streams optimally and avoid tuning the weights of different streams by hand, which is a difficult and expensive process, we use homoscedastic uncertainty as a basis to weigh losses of different streams automatically.


The main contributions of this paper are summarized as follows:

\begin{itemize}

\item
We introduce a semantic alignment loss to align the content of the features from the same category examples and compare relations among the elements inside an image for few-shot learning.

\item
We propose to maximize local and global mutual information, which allows for representations that contain locally-consistent and class-shared information across structural locations in an image.

\item
Homoscedastic uncertainty is introduced to learn the weights of different streams automatically, which can balance the streams optically and take full advantage of stream combination.

\item
Extensive experiments and analysis demonstrates that the proposed method is capable of comparing relations with semantic alignment strategies, and achieves state-of-the-art performance.

\end{itemize}

\section{Related Work}

\subsection{Few-shot learning}

Recent years have witnessed a surge of work on the few-shot learning task.
We briefly review the three main branches as follows.


\textbf{Matching network based methods} 
are based on the idea of matching, \textit{i.e.}, the samples belonging to the same category are more similar than the samples belonging to different categories in a feature space.
Embedding (mapping the samples to a feature space) and metric learning (measure the similarity) are two key steps.
Koch \textit{et al.} \cite{siamese_ICMLW15} learned image representations with siamese neural network to minimize the distance for similar samples and maximize for distinct ones.
Inspired from the ideas of deep neural features based metric learning and external memory augmented neural networks \cite{MANN_ICML16}, where the metric learning was used to provide good representation for the memory,
Vinyals \textit{et al.} \cite{Matching_Networks_NIPS16} proposed to learn a matching network to map labelled and unlabelled examples to their labels, where the output for a new class was described as a linear combination of the labels in the support set with an attention mechanism.
As an extension of matching networks, Prototypical networks \cite{Prototypical_NIPS17} learned a metric space in which classification can be performed by computing distances to prototype representations of each class.

For better metric learning ability, Sung \textit{et al.} \cite{Compare_CVPR18} added a relation module to compute the relation score between query images and the examples in the support set.
Oreshkin \textit{et al.} \cite{adaptive_metric_NIPS18} employed a task dependent adaptive metric for improved few-shot learning.
Instead of pairwise concatenation, Huang \textit{et al.} \cite{huang2019compare} utilized pairwise bilinear pooling to extract the second-order features for the pair of query images and support set images.
Hao \textit{et al.} \cite{SAML_ICCV19} calculated the similarity of each local region pairs of the query image and the support images, and used the similarity to reweight the pairs for comparison.
Besides convolutional neural networks and recurrent neural networks, Guo \textit{et al.} \cite{graph_matching_ECCV18} proposed Neural Graph Matching Networks, which jointly learned a graph generator and a graph matching metric function end-to-end.

For better feature learning ability, Li \textit{et al.} \cite{Local_Descriptor_CVPR19} replaced the image-level feature based measure with a local descriptor based measure, which was conducted online via a k-nearest neighbor search over the deep local descriptors.
Meanwhile, Lifchitz \textit{et al.} \cite{Dense_Classification_CVPR19} proposed dense classification over feature maps to take full advantage of the local activations and spatial information.
Li \textit{et al.} \cite{Task_Relevant_CVPR19} introduced a Category Traversal Module to traverse across the entire support set at once, identifying task-relevant features based on both intra-class commonality and inter-class uniqueness in the feature space.

\textbf{Meta-learner optimization based methods} are mainly derived from the idea of learning to learn or meta-learning. The models are learned at two levels, \textit{i.e.}, learning within each task, while accumulating knowledge between tasks. There are two optimizations, the learner, which adapts to new tasks, and the meta-learner, which trains the learner.
Santoro \textit{et al.} \cite{MANN_ICML16} proposed a memory-augmented neural network, which was trained to learn how to store and retrieve memories to use for each classification task.
There is a series of extensions based on memory-augmented neural network.
Shankar \textit{et al.} \cite{shankar2017labelMANN} organized the memory with the discrete class label as the primary key unlike the previous key being a real vector derived from the input.
Mureja \textit{et al.} \cite{mureja2017featurelabelMN} explicitly split the external memory into feature and label memories.
Ramalho \textit{et al.} \cite{adaptive_posterior_ICLR19} introduced adaptive posterior learning to approximate probability distributions by remembering the most challenging observations it had encountered.

Bertinetto \textit{et al.} \cite{bertinetto2016learning} constructed meta-learner and learner as two networks. The meta-learner network is trained to predict the parameters of the learner network.
Ravi \textit{et al.} \cite{Optimization_ICLR17} proposed an LSTM-based meta-learner model to learn both a good initialization and a parameter updating mechanism for the learner network.
Finn \textit{et al.} \cite{MAML_ICML17} trained the meta-learner to find an initialization that can be quickly adapted to a new task, via a few gradient steps.
Since the initial model of a meta-learner could be too biased towards existing tasks to adapt to new tasks, Jamal \textit{et al.} \cite{TAML_CVPR19} proposed an entropy-based approach that meta-learned an unbiased initial model with the largest uncertainty over the output labels.
Instead of forcibly sharing an initialization between tasks, Baik \textit{et al.} \cite{learn_forget_19} employed task-dependent layer-wise attenuation, which could dynamically control how much of prior knowledge each layer would exploit for a given task.
Li \textit{et al.} \cite{LGM_ICML19} learned to generate matching networks by learning transferable prior knowledge across tasks and directly producing network parameters for similar unseen tasks.

\textbf{Data augmentation based methods} try to increase the number of training set for better few-shot learning performance.
Xian \textit{et al.} \cite{f_VAEGAN_CVPR19} developed a conditional generative model that combined the strength of VAE and GANs, in addition, via an unconditional discriminator, to learn the marginal feature distribution of unlabeled images.
Zhang \textit{et al.} \cite{Hallucination_CVPR19} presented two light-weight data hallucination strategies for few-shot learning. Instead of GANs, they leveraged saliency network to obtain foreground-background pairs.
Chu \textit{et al.} \cite{patch_sampler_CVPR19} proposed a sampling method that decorrrelated an image based on maximum entropy reinforcement learning, and extracted varying sequences of patches on every forward-pass.
Self-supervised rotation was used as an auxiliary task in \cite{selfsupervision_ICCV19} to learn richer visual representations.
There are also some methods based on transfer learning \cite{Model_Regression_ECCV16, meta_transfer_CVPR19, knowledge_transfer_CVPR19} where knowledge learned from large enough sample sets are transferred to few samples.

The three branches can be combined in one method.
Zhang \textit{et al.} \cite{WACV2019_powernorm} utilized second-order statistics with power normalization and permutation-based data augmentation to learn the similarity for few-shot learning.
Wertheimer \textit{et al.} \cite{CVPR2019_localization} proposed batch folding, few-shot localization and covariance pooling for long-tailed class distributions with bounding box annotations.
Different from \cite{WACV2019_powernorm, CVPR2019_localization}, which used second-order statistics rather than first-order statistics to expand feature space, we use the first-order static appearance information, the second-order correlation information, and the local-global mutual information to constrain the model to learn consistent and intrinsic features to address content misalignment for few-shot learning.
In our relation stream, we enforce the images belonging to the same class to share a common correlation matrix across spatial positions as well as across channels. Both the relation among spatial local features and the relation among channel style features are taken into account, while other works only considered the first one.
We do not use any version of power normalization, bounding box annotations or permutation augmentations in our method.
Furthermore, the consistency relationship among features and the multi-stream fusion in our model are optimized with a mutual information stream and uncertainty-based weighting mechanism.


\subsection{Uncertainty-based learning} 

Given a model and an input, the uncertainty corresponds to the confidence level of the outcome predicted by the model.
In data analysis, it is essential to not only provide a good model but also an uncertainty estimate of the conclusions.
It can also be used to improve the robustness and generalization ability of the model.
Wang \textit{et al.} \cite{wang2017selective} explicitly formulated a propagation uncertainty term to guide the selection of the ambiguous frames for user annotation in video cutout, which reduced manual effort and improved segmentation performance.
Khan \textit{et al.} \cite{khan2019striking} incorporated both class and sample-level uncertainty estimates to re-adjust the learned boundaries of the classifier for learning unbiased models on imbalanced datasets.
Mustafa \textit{et al.} \cite{mustafa2020deeply} decreased the uncertainty of the model against attacks by forcing the features of each class to be maximally separated from the polytopes of other classes.

In Bayesian modeling, uncertainty can be categorized into two types one can model: epistemic uncertainty, which accounts for uncertainty in the model parameters, and aleatoric uncertainty, which captures noise inherent in the observations \cite{der2009aleatory}. Aleatoric uncertainty can be further categorized into homoscedastic uncertainty, which stays constant for different inputs, and heteroscedastic uncertainty.
Kendall \textit{et al.} \cite{kendall2017uncertainties} presented a Bayesian deep learning framework combining input-dependent heteroscedastic uncertainty with epistemic uncertainty to improve model's robustness.
For multi-task learning, Kendall \textit{et al.} \cite{kendall2018multi} utilized homoscedastic uncertainty to weigh losses from different tasks, in order to improve the performance of the model on each task.
In our paper, we use homoscedastic uncertainty to automatically aggregate the outputs from multiple streams, avoiding tuning the weights by hand. The tasks of the streams could be different or the same. The final goal is to improve the performance of the output after aggregation. We extend the uncertainty-based weighting strategy from multi-task learning scenario to multi-stream fusion problem, which stimulates its use for automatic weight learning and generalizes its application area.

\section{Proposed Method}

In this section, we give a detailed introduction to the proposed model as shown in Figure \ref{fig:framework}.
Besides an appearance stream, we additionally introduce a relation stream and a mutual information (MI) stream to align content and compare relations. These three streams are complementarily combined, resulting in a model robust to content misalignment.

\subsection{Network Architecture} 

There are three major components in the proposed network, the embedding module $f_\varphi$, the metric learning module $g_\phi$ and the fusion module.
We take a commonly used four-layer convolutional network as the embedding module by default. With the same setting of \cite{Compare_CVPR18}, it contains four convolutional blocks. The first two blocks have an architecture of $convolution\ (3 \times 3, channels=64) \to max\ pooling\ (2 \times 2, stride=2) \to batch\ normalization \to ReLU$, where the first digit in parenthesis indicates the kernel size. The last two blocks have an architecture of $convolution\ (3 \times 3, channels=64) \to batch\ normalization \to ReLU$. This embedding network is named $Conv$-$64F$, since there are 64 filters in each convolutional block.
More analysis with a deeper embedding network based on ResNet is shown in the experiment.
The metric learning module includes an appearance stream $g_\phi^s$, a relation stream $g_\phi^r$ and a mutual information stream $g_\phi^{MI}$, which we will introduce in the following subsections.

For few-shot learning, there is a support set of example images $\mathcal{S} =\{({x_i},{y_i})\}_{i = 1}^m$ ($m=K\times C$) which contains $K$ labeled samples for each class out of $C$ classes. It can also be called $C$-way $K$-shot learning.
The task is to predict the classes of the images in a query set $\mathcal{Q}=\{({x_j},{y_j})\}_{j = 1}^n$ according to $\mathcal{S}$.
Episodic training mechanism \cite{MANN_ICML16, Matching_Networks_NIPS16} simulates the few-shot setting to train the network.
At each training iteration, $N$ sets of $\{ \mathcal{S}$, $\mathcal{Q \}}$ are randomly generated from the entire training set, where $N$ denotes the batch size.
The images from the support set and the query set are fed forward into the embedding module for feature extraction. Then the metric learning module compares the feature of the query image with the representations corresponding to $C$ classes based on the $K$-shot examples to generate a set of matching scores.
Finally, the fusion module aggregated these scores to form a global matching score.
Each element of the matching score represents the similarity between the query image and the example image among $C$ classes, which can predict a probability score for classification. The label is predicted by choosing the category with the maximum probability value.

\subsection{Appearance Stream}

\begin{figure}[!t]
\centering
\includegraphics[width=\columnwidth]{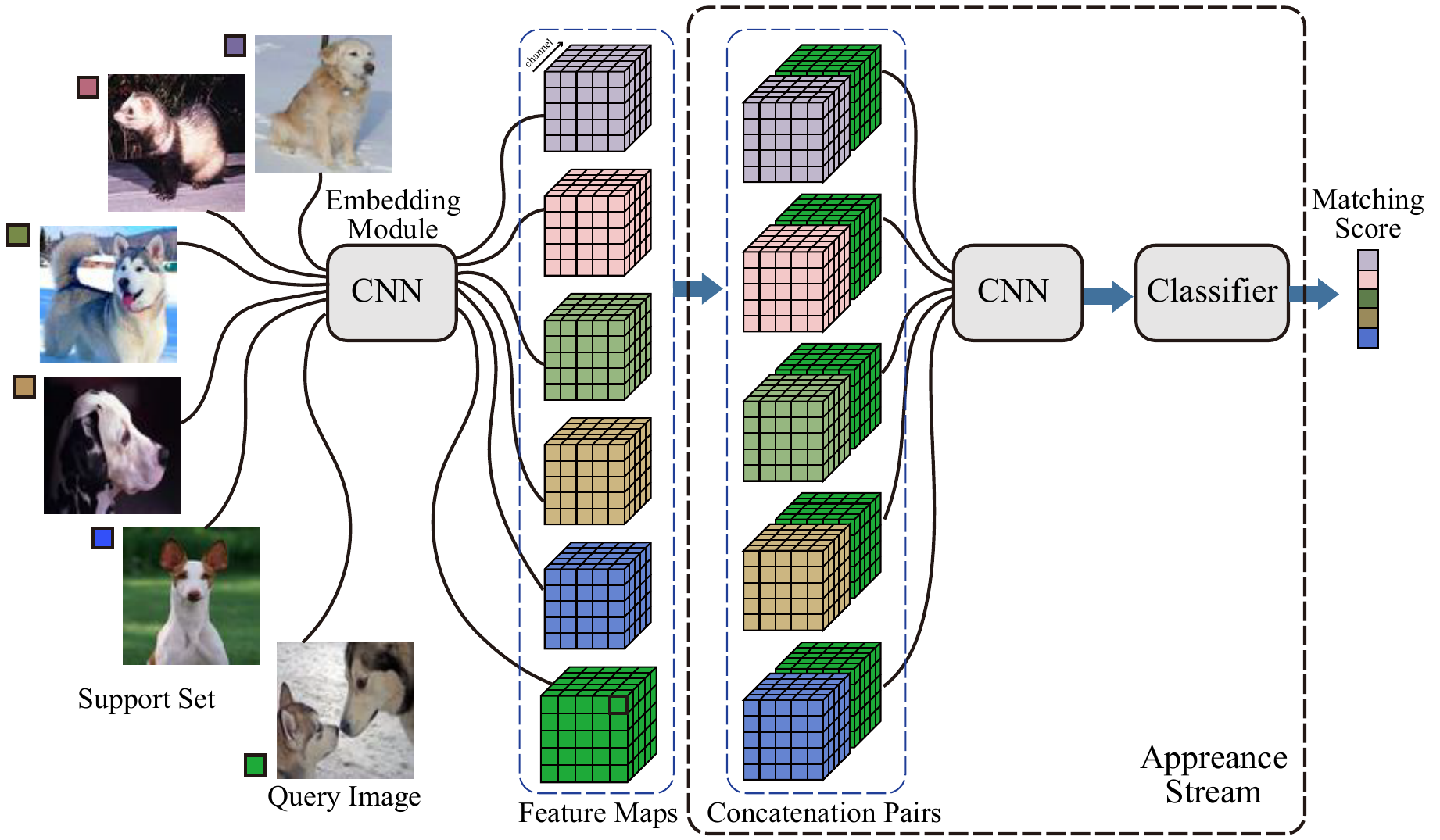}
\caption{The appearance stream. The feature maps generated from the embedding module corresponding to the support images and the query image are concatenated to form feature pairs. Then the feature pairs are fed forward into a CNN and a classifier for matching score prediction.}
\label{fig:appearance}
\end{figure}

\begin{figure*}[!t]
\centering
\includegraphics[width=1.8\columnwidth]{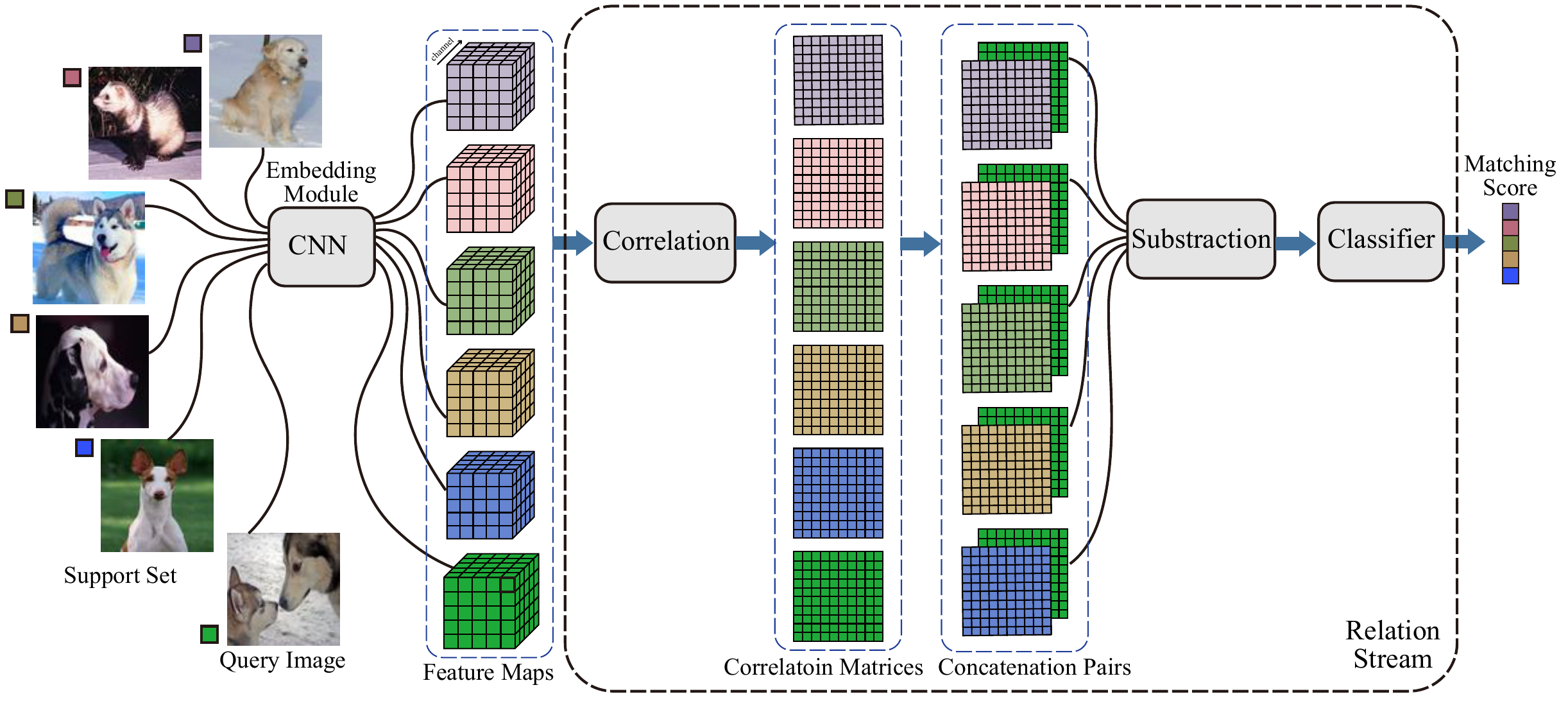}
\caption{The relation stream. Correlation statistics of the embeddings are computed to characterize the relation information inside an image. Then the correlation matrices of the support images and the query image are concatenated to compute distance. At last, a classifier is trained to predict a set of matching scores based on the distances between the support images and the query image.}
\label{fig:relation}
\end{figure*}

The appearance stream as shown in Figure \ref{fig:appearance} is retained from the existing few-shot learning work \cite{Compare_CVPR18}.
It replaced non-parametric similarity measure, such as Euclidean distance and cosine similarity, with a network-learned similarity measure.
Although it was called RelationNet in the original paper, it focused on comparing local appearance similarity between two images, while discarding the relation among the elements inside an image, which is different from our relation stream.
We keep using this stream for appearance information comparison and using the same criteria to measure the pairwise similarities in other streams for convenience and fairness.

Let $f_\varphi(x_i)$ and $f_\varphi(x_j)$ denote the feature maps generated by the embedding module with the input of example image $x_i$ and query image $x_j$ respectively. To determine whether $x_i$ and $x_j$ are from matching class or not, the appearance stream $g_\phi^s$ predicts a matching score $p_{i,j}^s$ based on the concatenated representation of $f_\varphi(x_i)$ and $f_\varphi(x_j)$. The architecture of $g_\phi^s$ is: two convolutional blocks with the same architecture of the first two blocks in $f_\varphi$, followed by $fc \ (8) \to ReLU \to fc \ (1) \to sigmoid$, where the digit in parenthesis indicates the filter numbers.

It should be noted that RelationNet is just one of the choices for the appearance stream. More analyses based on different appearance streams and similarity measures are introduced and shown in Section \ref{section:implementation} and \ref{section:baselines}.


\subsection{Relation Stream}   

In an ideal case, for the samples from the same class, the representations are expected to be consistent and share common semantics. However, in practice, it is hard to learn the high-level complex concept since there is a wide variation and inevitable noise in the samples, especially with only a few examples.
In order to learn semantic concepts robust to variation and noises, we propose to use correlation for feature representation and relation comparison. The proposed relation stream is shown in Figure \ref{fig:relation}. The stream uses multiplicative interactions among local features inside an image to represent the relation information, and measures the image-pairs similarity based on it.


Given $f_\varphi(x_i)$, $f_\varphi(x_j)$ $ \in {\mathbb{R} ^{C \times H \times W}}$, where $C$, $H$, $W$ denote the number of channels, height and width of the feature maps respectively, the feature map can be seen as two feature sets from different aspects. One is the spatial feature set, the other is the channel feature set.
Let $F_i$, $F_j$ $ \in {\mathbb{R} ^{C \times D}}$ ($D=H\times W$) denote the reshaped feature maps.
The columns of $F_i$, $F_j$ are spatial local features. The rows are channel style features.
In order to take both the relationship between spatial local features and the relationship between channel style features into account, two correlation matrices are calculated:


\begin{eqnarray}
corr_D({F_i}) &=& norm{({F_i})^T}norm({F_i}) \in {\mathbb{R} ^{D \times D}}\\
corr_C({F_i}) &=& norm({F_i})norm{({F_i})^T} \in {\mathbb{R} ^{C \times C}}\\
norm({F_i}) &=& {\raise0.7ex\hbox{${\frac{{{F_i} - {\mu _i}}}{{{\sigma _i}}}}$} \!\mathord{\left/
 {\vphantom {{\frac{{{F_i} - {\mu _i}}}{{{\sigma _i}}}} {\left\| {\frac{{{F_i} - {\mu _i}}}{{{\sigma _i}}}} \right\|}}}\right.\kern-\nulldelimiterspace}
\!\lower0.7ex\hbox{${\left\| {\frac{{{F_i} - {\mu _i}}}{{{\sigma _i}}}} \right\|}$}}
\label{equation:correlation}
\end{eqnarray}
where $norm({F_i})$ represents the normalization of ${F_i}$ by subtracting the mean ${\mu _i}$ and dividing the standard-deviation ${\sigma _i}$ of the elements. ${\left\| \cdot \right\|}$ is the operation to calculate magnitude. Superscript $(\cdot)^T$ stands for transposition.

Note that the correlation matrix $corr_C({F_i})$ is a Gram matrix that usually used in style transfer tasks \cite{style_transfer_NIPS15, style_transfer_CVPR16}. Covariance matrix has also been used to align the source and target feature maps in domain adaptation \cite{correlation_alignment_ECCVW16, correlation_alignment_ICLR18}. We align both the statistics of positional information and the statistics of style information for few-shot learning by predicting matching score based on the distance of the correlation matrices as following:

\begin{eqnarray}
{D_{i,j}^r} &=& \left\| {corr({F_i}) - corr({F_j})} \right\| \\
{p_{i,j}^r }   &=& g_\phi^r (D_{i,j}^r)
\label{equation:corrDiff}
\end{eqnarray}
where $g_\phi^r$ represents a fully-connected layer followed by a sigmoid function. The subscripts of $corr_D$ and $corr_C$ are omitted for clarity.



\subsection{Mutual Information Stream}   

\begin{figure*}[!t]
\centering
\includegraphics[width=1.8\columnwidth]{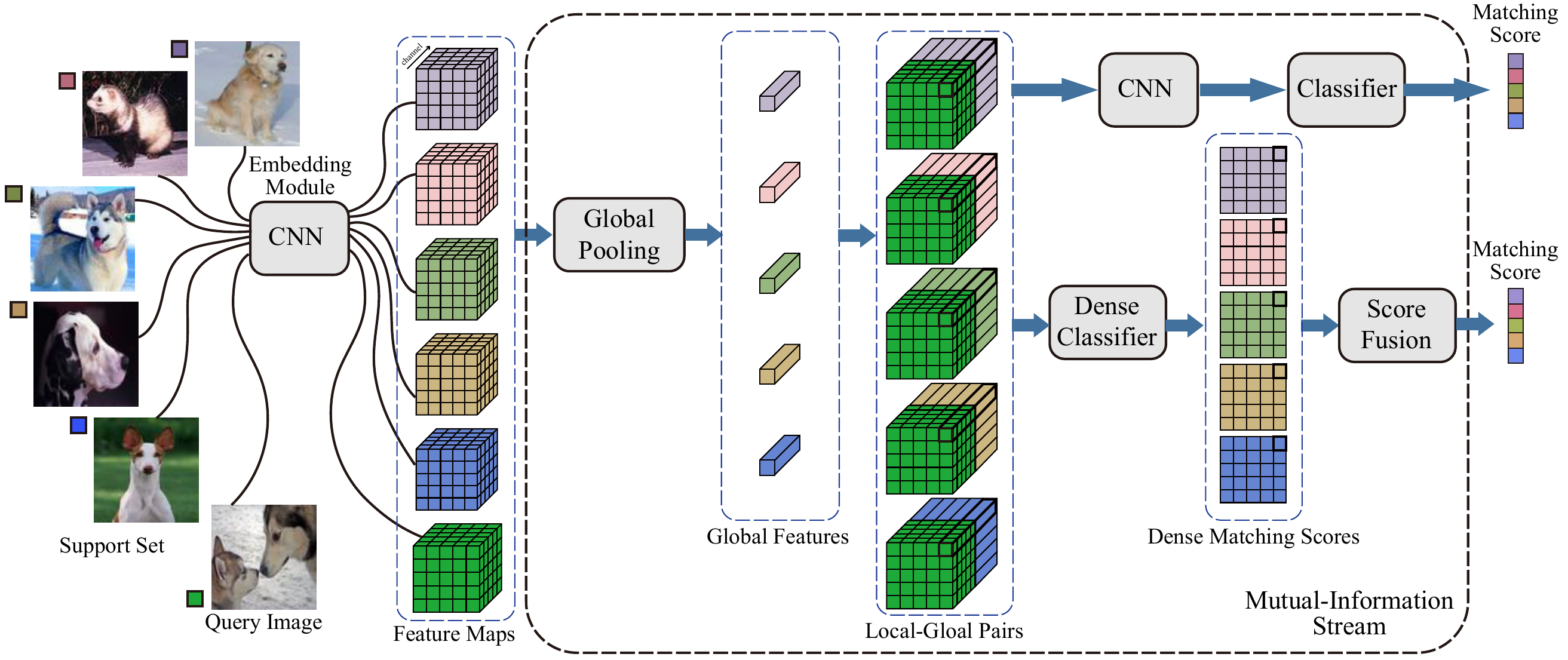}
\caption{The local-global mutual information stream. Global features of the support images are concatenated with each local features of the query image to obtain a local-global pair-wise representation. Matching scores are predicted based on the mutual information between the local-global representation pairs.}
\label{fig:MI}
\end{figure*}

Mutual information (MI) describes the association between two variables. Instead of only comparing feature statistics between the support images and the query image, we propose to maximize the mutual information between the images belonging to the same class while minimizing the mutual information between the images from different classes.
To learn locally-consistent and class-shared representations across structural locations in the image, we propose to optimize the mutual information between query's local features and examples' global description. 
The proposed local-global MI stream helps the network to learn and compare essential representations regardless of content misalignment and noise. It is complementary to the relation stream.
Figure \ref{fig:MI} illustrates the structure of MI stream.


For a support set image $x_i$, its feature map encoded by the embedding module $f_\varphi(x_i)$ is summarized into a global feature ${E_\varphi }({x_i})$ by a global pooling layer. We also have tried other architectures such as using a convolutional sub-network to summarize the feature map. However, experiments show that global max pooling performs superior to average pooling and other architectures.
For the query image $x_j$, its feature map encoded by the embedding module $f_\varphi(x_j)$ can be seen as a set of spatial local features $\{ f_\varphi ^{(d)}({x_j})\} _{d = 1}^{H \times W}$, index by $d$.
We concatenate each $f_\varphi ^{(d)}({x_j})$ with ${E_\varphi }({x_i})$ to obtain a new feature map to represent local-global pairs, as shown in Figure \ref{fig:MI}. The local-global mutual information $\mathcal{I}$ is estimated with a convolutional sub-network and two fully-connected layers denoted as $g_\phi^{MI}$. We use the estimated local-global mutual information to measure the similarity between two images, \textit{i.e.}, the matching score $p_{i,j}^{MI}$:

\begin{eqnarray}
{p_{i,j}^{MI} }   &=& \mathcal{I}(\{ f_\varphi ^{(d)}({x_j})\} _{d = 1}^{H \times W},{E_\varphi }({x_i})) \\ \nonumber
&=& {g_\phi^{MI}(\{ f_\varphi ^{(d)}({x_j})\} _{d = 1}^{H \times W},{E_\varphi }({x_i}))}
\label{equation:MI}
\end{eqnarray}

We can also use a dense classifier to estimate the mutual information, \textit{i.e.}, the matching scores on each local-global representation pair separately. Then the dense matching scores are fused to obtain an image-level aggregated matching score.

\begin{eqnarray}
{p_{i,j,d}^{MI} } &=& \mathcal{I}(f_\varphi ^{(d)}({x_j}),{E_\varphi }({x_i})) \\  \nonumber
&=& g_\phi^{MI}(f_\varphi ^{(d)}({x_j}),{E_\varphi }({x_i})) \\
{p_{i,j}^{MI} } &=& \frac{1}{{H \times W}}\sum\limits_{d = 1}^{H \times W} {p_{i,j,d}^{MI} }
\label{equation:MIdense}
\end{eqnarray}

\subsection{Full Objective of the Network}

Both Mean square error (MSE) and cross-entropy loss (CEL) can be used to train the model.
Combining the aforementioned streams, the full objective for training the matching network is as follows:

\begin{eqnarray}
L = {w_a}{L_a} + w{}_r{L_r} + w{}_{MI}{L_{MI}}
\label{equation:objective}
\end{eqnarray}
where ${w_a}$, ${w_r}$, and ${w_{MI}}$ are the weights for the loss ${L_a}$, ${L_r}$, and ${L_{MI}}$ of different streams, which can be computed as:
\begin{eqnarray}
{L_{a,r,MI}} = \left\{ {\begin{array}{*{20}{c}}
{\frac{1}{2}\sum\limits_{j = 1}^n {\sum\limits_{i = 1}^m {{{({y_{i,j}} - p_{i,j}^{a,r,MI})}^2},\ MSE} } }\\
{ - \sum\limits_{j = 1}^n {\sum\limits_{i = 1}^m {{y_{i,j}}\log (p_{i,j}^{a,r,MI}),\ CEL} } }
\end{array}} \right.
\label{equation:loss}
\end{eqnarray}
where the supervision $y_{i,j}$ is 1 for the image pairs belonging to the same class and 0 for mismatched pairs.
\begin{eqnarray}
{y_{i,j}} = \left\{ {\begin{array}{*{20}{c}}
{1,{y_i} = {y_j}}\\
{0,{y_i} \ne {y_j}}
\end{array}} \right.
\label{equation:y}
\end{eqnarray}

\subsection{Weight Learning with Homoscedastic Uncertainty}

Manual tuning the weights of different streams, \textit{i.e.}, ${w_a}$, ${w_r}$, and ${w_{MI}}$, is time-consuming. It is preferable to learn the optimal weights automatically with the training process of the whole network.
It was first analyzed in multi-task learning \cite{kendall2018multi} that homoscedastic uncertainty could be used to combine multiple loss functions. We apply the idea of probabilistic modeling to fusing the three streams in our proposed model automatically.
Homoscedastic uncertainty is a kind of aleatoric uncertainty which captures noise inherent in the model's outcome and is independent of the input data. In our multi-stream setting, it varies among different streams, which reflects the relative confidence among streams.
It also depends on the measurement scale of each stream. Hence it can be used as a basis for weighting the outcomes in a multi-stream fusion problem.
According to the theoretical analysis and formula deduction in \cite{kendall2018multi}, the likelihood as a Gaussian of a regression task with MSE loss is:

\begin{eqnarray}
p(y|h(x)) &=& {\rm N}(h(x),\sigma _{}^2)\\
\log p(y|h(x)) &\propto&  - \frac{1}{{2\sigma _{}^2}}{(y - h(x))^2} - \log \sigma \\ \nonumber
 &=&  - \frac{1}{{\sigma _{}^2}}L_{MSE} - \log \sigma
\label{equation:regression}
\end{eqnarray}
where $y$ represents the ground-truth supervision of a stream, $h(x)$ represents the output of the stream with input $x$ ($h$ contains feature embedding module $f$ and stream-specific part $g$).
$\sigma$ is the observation noise scalar.

The likelihood of a classification task with CEL loss can be written as a scaled version:

\begin{eqnarray}
&p(y|h(x)) = Soft\max (\frac{1}{{\sigma ^2}}h(x)) \\
&\log p(y = c|h(x)) = \frac{1}{{{\sigma ^2}}}h_{}^c(x) - \log \sum\limits_i {\exp (\frac{1}{{{\sigma ^2}}}h_{}^i(x))}  \\ \nonumber
&\approx  - \frac{1}{{{\sigma ^2}}}{L_{CEL}} - \log \sigma
\label{equation:classification}
\end{eqnarray}
where $\sigma ^2$, which often referred to as $temperature$, is used to scale the input.


We can derive a joint loss of the three streams with homoscedastic uncertainty based on maximizing the log likelihood of the streams:

\begin{eqnarray}
L =  &- \log p({y^a}|{h^a}(x)) - \log p({y^r}|{h^r}(x)) \\  \nonumber &- \log p({y^{MI}}|{h^{MI}}(x)) \\  \nonumber
\propto &\frac{1}{{\sigma _a^2}}{L_a} + \frac{1}{{\sigma _r^2}}{L_r} + \frac{1}{{\sigma _{MI}^2}}{L_{MI}} \\  \nonumber
&+ \log {\sigma _a} + \log {\sigma _r} + \log {\sigma _{MI}}
\label{equation:learnable_w}
\end{eqnarray}

It should be noted that, regardless of whether the streams correspond to the same task or different tasks, the joint loss can be formulated as above. It applies to our multi-stream fusion objective as shown in Equation \ref{equation:objective} and can be seen as learning the relative weights of the losses for each stream.
Large $\sigma$ decreases the influence of the corresponding stream's $L$, while small $\sigma$ increases its influence. The last three terms are the regularization item, which penalise too large $\sigma$ and guarantee that the weights will not converge to zero. Therefore the weights for streams can be optimized automatically with the whole network. For more stable computation, log variance $s \buildrel \Delta \over = \log {\sigma ^2}$ is used to avoid any division by zero. The objective is written as:

\begin{eqnarray}
L &= {e^{ - {s_a}}}{L_a} + {e^{ - {s_r}}}{L_r} + {e^{ - {s_{MI}}}}{L_{MI}} \\ \nonumber &+ \frac{1}{2}({s_a} + {s_r} + {s_{MI}})
\label{equation:learnable_s}
\end{eqnarray}
\textit{i.e.}, $e^{ - {s_a}}$, $e^{- {s_r}}$, and $e^{- {s_MI}}$ are the learnable weights for the loss ${L_a}$, ${L_r}$, and ${L_{MI}}$ of different streams.

\section{Experiments}

\subsection{Datasets}

We conduct our experiments on six publicly available few-shot learning datasets with two tasks, \textit{i.e.}, 5-way 1-shot learning and 5-way 5-shot learning.

\textbf{miniImageNet} \cite{Matching_Networks_NIPS16} is a subset of ImageNet \cite{russakovsky2015imagenet}, which contains 100 classes with 600 images in each class. The spatial resolution of the images is $84 \times 84$ as default setting. Following the same data splits setting of \cite{Optimization_ICLR17}, 64, 16 and 20 classes are taken for training, validation and testing respectively. For $C$-way $K$-shot learning, besides the $K$ examples for each class, there are 15 and 10 query images for 5-way 1-shot and 5-way 5-shot learning respectively in each training episode. For testing, accuracy averaged over 600 randomly generated episodes is used to measure the performance.

\textbf{Omniglot} \cite{lake2011one} consists of 1623 character classes from 50 alphabets. Each class contains 20 samples drawn by 20 people. 
The spatial resolution of the input images is $24 \times 24$. 1200 classes are used for training, and the remaining 423 classes are used for testing.
We follow the common data augmentation setting and training setting of \cite{Compare_CVPR18}.
For $C$-way $K$-shot learning, besides the $K$ examples for each class, there are 19, 15, 10, and 5 query images for 5-way 1-shot, 5-way 5-shot, 20-way 1-shot, and 20-way 5-shot learning respectively in each training episode. When testing, accuracy averaged over 1000 randomly generated episodes is used to measure the performance.

\textbf{CUB} \cite{CUBdataset} is initially designed for fine-grained classification, which is challenging for few-shot learning since the species are not substantially different from each other. There are 11,788 images of birds over 200 species. Following the commonly used data splitting setting, we randomly sampled 100 species for training, 50 species for validation, and 50 species for testing. We crop the images with the provided bounding box as a pre-processing \cite{triantafillou2017few}. The other settings are the same as those on miniImageNet dataset.

\textbf{tieredImageNet} \cite{ren2018meta} is a larger subset of ImageNet composed of 608 classes rather than 100 for miniImageNet. These classes belong to 34 higher-level categories. To ensure that the training classes are distinct from the testing classes, data splitting is performed at category-level: 20 categories for training, 6 for validation and 8 for testing. The numbers of classes for training, validation and testing are 351, 97 and 160 respectively. The spatial resolution of the input images is 84 $\times$ 84. Following \cite{ye2020few}, accuracy is averaged over 10,000 sampled tasks for more trustworthy evaluation.

\textbf{Flower-102 and Food-101} \cite{nilsback2008automated, bossard2014food} are two other fine-grained datasets like CUB. There are 102 classes of flowers in Flower-102, each containing 40-258 images. And Food-101 is composed of 101 food classes with 1000 images in each class. For both datasets, following \cite{zhang2020few}, we randomly select 80 classes for training and use the remaining classes for testing. The other settings are the same as those on tieredImageNet.

\begin{table*}[!t]
\caption{Performance analysis of the relation stream, local-global mutual information stream, and stream fusion with different weights. The experiments are carried out on the miniImageNet dataset with 5-way 1-shot setting.}
\label{table:ablation}
\centering
\begin{tabular}{l|c|c c c c c|c}
\hline
\hline
\multicolumn{2}{c|}{Setting} & $appearance$ & $corr_C$ & $corr_D$ & $MI$ & $MIdense$ & $acc (\%)$  \\
\hline
\multicolumn{2}{c|}{$appearance$ only} & 1 & 0 & 0 & 0 & 0 & 49.80+-0.82  \\
\multicolumn{2}{c|}{$corr_C$ only}   & 0 & 1 & 0 & 0 & 0 & 49.67+-0.83  \\
\multicolumn{2}{c|}{$corr_D$ only}   & 0 & 0 & 1 & 0 & 0 & 49.63+-0.81  \\
\multicolumn{2}{c|}{$MI$ only}       & 0 & 0 & 0 & 1 & 0 & 49.02+-0.86  \\
\multicolumn{2}{c|}{$MIdense$ only}  & 0 & 0 & 0 & 0 & 1 & 48.03+-0.86  \\
\hline
\multirow{10}*{\textit{manual tuning}} & \multirow{4}*{$+corr$}  & 0 & 2 & 1 & 0 & 0 & 51.81+-0.86  \\
~  & ~   & 2 & 1 & 0 & 0 & 0 & 52.00+-0.86  \\
~  & ~   & 2 & 0 & 1 & 0 & 0 & 51.03+-0.86  \\
~  & ~   & 4 & 2 & 1 & 0 & 0 & 53.06+-0.88  \\
\cline{2-8}
~  & \multirow{3}*{$+MI$}  & 1 & 0 & 0 & 1 & 0 & 52.22+-0.87  \\
~  & ~   & 1 & 0 & 0 & 0 & 1 & 52.23+-0.86  \\
~  & ~   & 1 & 0 & 0 & 1 & 1 & 52.23+-0.87  \\
\cline{2-8}
~  & $corr+MI$   & 0 & 2 & 1 & 4 & 0 & 52.67+-0.88  \\
\cline{2-8}
~  & 3 streams  & 4 & 2 & 1 & 2 & 0 & 53.30+-0.88  \\
\hline
\multicolumn{2}{c|}{\multirow{4}*{\textit{weight learning}}}  & \checkmark & \checkmark & \checkmark &  &  & 53.03+-0.86  \\
\multicolumn{2}{c|}{~}   & \checkmark &  &  & \checkmark &  & 52.25+-0.87  \\
\multicolumn{2}{c|}{~}   &  & \checkmark & \checkmark & \checkmark &  & 52.85+-0.88  \\
\multicolumn{2}{c|}{~}   & \checkmark & \checkmark & \checkmark & \checkmark & & \textbf{53.33}+-0.88  \\
\hline
\hline
\end{tabular}
\end{table*}

\subsection{Implementation Details}  \label{section:implementation}

By default, the 4-layer convolutional network $Conv$-$64F$ is used as the embedding module and the network-learned similarity measure like RelationNet is used. In this case, except for the MI stream, the networks used for miniImageNet, Omniglot, and CUB datasets are the same. To estimate the local-global mutual information, two convolutional blocks with the same architecture of the first two blocks in $f_\varphi$ followed by two fully-connected layers with the same architecture of $g_\phi$ are used for 84$\times$84 input.
Since the spatial resolution of Omniglot is small, which is $24\times24$, we remove the last pooling layer for this dataset.
All the models are trained end-to-end from scratch with random initialization.
Adam is used for optimization. The initial learning rate is $10^{-3}$ and reduced with a fixed decay factor 2 every 100,000 epochs.
In our experiments, MSE and CEL have similar performances. We report the results based on MSE as default.

Besides RelationNet with the 4-layer convolutional embedding module, we further demonstrate the usefulness and generalization ability of our method by applying it with different baseline models, \textit{e.g.}, SoSN \cite{WACV2019_powernorm} with the 4-layer embedding module, as well as ProtoNet \cite{Prototypical_NIPS17} and RelationNet \cite{Compare_CVPR18} with a ResNet-12 embedding module. They are based on different similarity measures, \textit{e.g.}, negative Euclidean distance for ProtoNet, and network-learned similarity for RelationNet and SoSN. For the proposed multi-stream model, we keep using the same similarity measure with the baseline model for convenience and fairness. The ResNet-12 embedding module has a same structure with the 12-layer residual network used in \cite{lee2019meta} and \cite{ye2020few}.
With a deep embedding module, pre-training on the meta-training set with classification task is suggested and widely used \cite{RusuRSVPOH19, Task_Relevant_CVPR19, ye2020few}.
We use the pre-trained ResNet-12 embedding module of \cite{ye2020few} without the last global average pooling layer for parameter initialization. The training and evaluation settings are consistent with the baseline model.

\subsection{Ablation Study}

\subsubsection{Correlation Matrices Comparison}


Firstly, we evaluate the performance of the proposed relation stream and add it to the existing appearance matching network \cite{Compare_CVPR18}.
The performance of the appearance stream reported in Table \ref{table:ablation} is reproduced by us with the same setting of \cite{Compare_CVPR18}.
$corr_D$ represents the relation stream based on optimizing the similarity of the correlation matrix between spatial local features. And $corr_C$ represents the relation stream based on optimizing the similarity of the correlation matrix between channel style features.
As we can see, both relation stream $corr_D$ and relation stream $corr_C$ have a similar performance with the commonly used appearance stream. Note that the relation stream focuses on the relation among the local features inside an image regardless of absolute value and position, which is robust to misalignment and noise. Furthermore, it can be a necessary complement to existing methods. The performance is significantly improved by combing the relation stream and the appearance stream together as shown in Table \ref{table:ablation}.
The experimental results demonstrate the effectiveness of the proposed relation stream, indicating that relation is an important factor in few-shot learning and it is complementary to the existing appearance comparison methods.

\subsubsection{Mutual Information Optimization}


\begin{figure*}[!t]
\centering
\includegraphics[width=1.8\columnwidth]{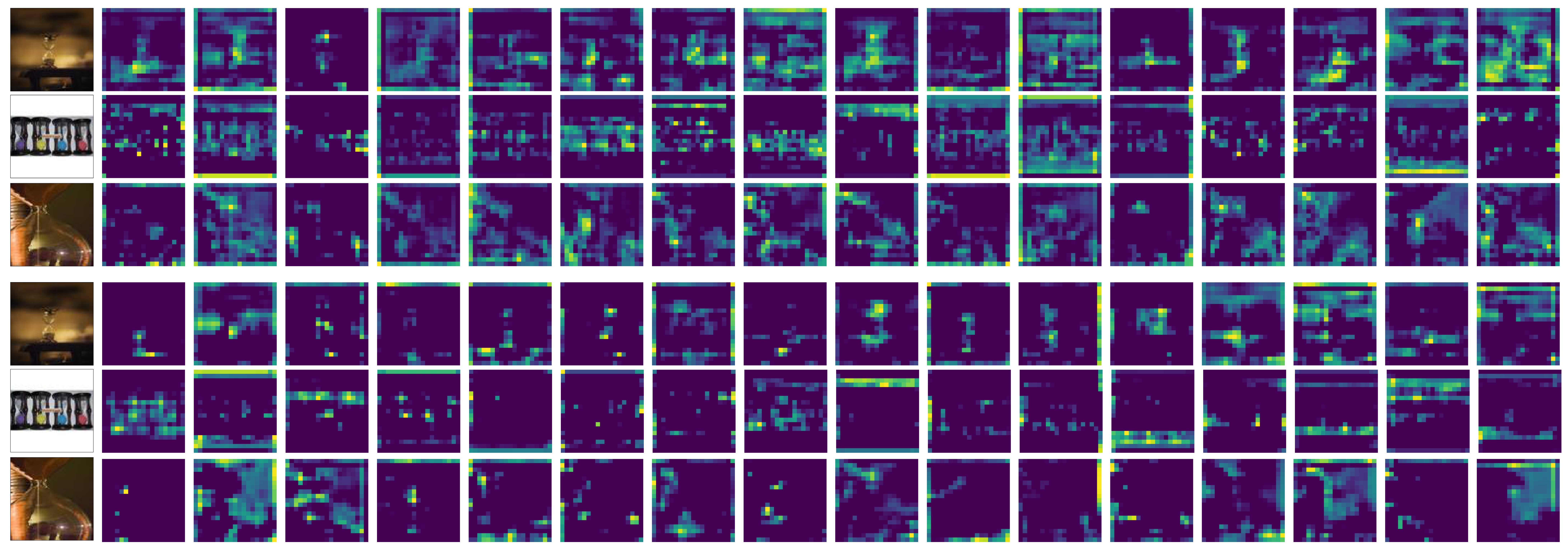}
\caption{Feature maps of the embedding module trained with and without the MI stream. The input are 3 image samples randomly chosen from the same class and the first 16 channels are shown. The feature maps in the first three rows are more locally-consistent compared with those in the last three row.}
\label{fig:featuremaps}
\end{figure*}

\begin{figure}[!t]
\centering
\includegraphics[width=0.8\columnwidth]{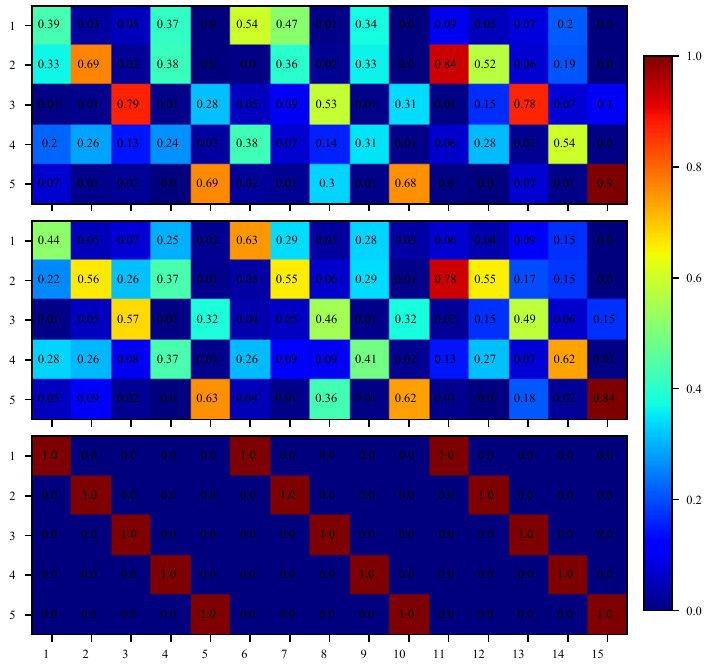}
\caption{Matching scores of the appearance stream, the proposed model and the ground-truth labels in a batch. Vertical axis and horizontal axis denote the class and sample indexes respectively.}
\label{fig:visualization}
\end{figure}

Then we evaluate the performance of the proposed local-global mutual information (MI) stream. The experimental results are also listed in Table \ref{table:ablation}. Simply optimizing the mutual information between the local patches of the query image and the global information of the examples in support set can achieve a performance only slightly lower than that of the appearance comparison method. However it can fundamentally improve the representation learning ability of the network. The MI stream encourages the network to prefer information that is shared across the samples belonging to the same class. It also allows for representations that contain locally-consistent information across structural locations which can deal with misalignment and noise effectively. Experimental results support the above hypothesis since regardless of combining the local-global MI stream with the existing appearance comparison method or with the proposed relation stream, the recognition accuracy can be improved significantly for few-shot learning. Since the two versions of the MI stream play the same role and share a similar performance, we no longer use the dense classifier in the following experiments.

To show that the learned representations are more locally-consistent, we visualize the feature maps of the embedding module trained with and without the MI stream. The input are 3 image samples randomly chosen from the same class and the first 16 channels are shown in Figure \ref{fig:featuremaps}. We use a sub-network to estimate the local-global MI and set the ground-truth MI for the images belonging to the same class and different classes to be 1 and 0 respectively. The MSE is decreased from 0.4367 to 0.0971 with the MI stream, which supports that applying these local-global pairs can maximize the MI of samples from the same class.

\subsubsection{Multi-Stream Fusion}


When manual tuning the weights, we evaluate the weights of \{1,2,4\} for each stream. Taking two streams fusion for example, there are five groups of weights, \textit{i.e.}, \{(4:1),(2:1),(1:1),(1:2),(1:4)\}. As the number of streams increases, the number of weight combinations grows. We only report the combination with the highest performance in Table \ref{table:ablation}.
Since the relation stream and the MI stream are designed to solve misalignment and noise from different perspectives, they are complementary to each other and the appearance stream.
As shown in Table \ref{table:ablation}, when the weights of the appearance stream, relation stream $corr_C$, relation stream $corr_D$, and MI stream are $\{4, 2, 1, 2 \}$, the recognition accuracy for 5-way 1-shot learning on miniImageNet is $53.30\%$. The performance is improved by $3.50\%$ with the proposed relation comparison method.
We visualize the matching scores of 15 samples in a batch for the appearance stream and the proposed model in Figure \ref{fig:visualization}.

In order to avoid tuning the weights of different streams by hand, which is a difficult and expensive process, we use homoscedastic uncertainty to weigh losses of different streams automatically. The result is listed in the last block of Table \ref{table:ablation}. With the automatical weight learning method, we only need to train the multi-stream model once, while the performance is comparable or even better than that of manual tuning with multiple times training.


\subsubsection{Combination with Different Baseline Models} \label{section:baselines}


\begin{table}[!t]
\caption{Combination with different baseline models. The experiments are carried out on the miniImageNet dataset with 5-way 1-shot setting.}
\label{table:baselines}
\centering
\begin{tabular}{l|c|c|c|c}
\hline
\hline
Model & Baseline & $+corr$ & $+MI$ & \makecell[l]{$+corr$ \\ $+MI$}  \\
\hline
RelationNet \cite{Compare_CVPR18} & 49.80 & 53.03 & 52.25 & 53.33 \\
SoSN \cite{WACV2019_powernorm} ($corr_C$) & 52.96 & 54.47 & 54.12 & 54.82 \\
\hline
RelationNet (ResNet-12) \cite{Compare_CVPR18} & 61.51 & 61.97 & 62.39 & 62.68 \\
ProtoNet (ResNet-12) \cite{Prototypical_NIPS17} & 62.39 & 62.74 & 63.19 & 63.47  \\
\hline
\hline
\end{tabular}
\end{table}

Besides the original RelationNet with the 4-layer CNN embedding module, we further apply our method with different baseline models to demonstrate its usefulness and generation ability. The baseline models are SoSN \cite{WACV2019_powernorm} with the 4-layer embedding module, as well as ProtoNet \cite{Prototypical_NIPS17} and RelationNet \cite{Compare_CVPR18} with the ResNet-12 embedding module as introduced in Section \ref{section:implementation}. The experimental results are listed in Table \ref{table:baselines}.
Since SoSN already used the second-order correlation across spatial positions which is called $corr_C$ in our paper, we add the relation stream with the correlation across channels, \textit{i.e.}, $corr_D$, and the MI stream to it. With the proposed method, the performance is improved by 3.53\% and 1.86\% for RelationNet and SoSN respectively based on 4-layer embedding module.
When using the deep pre-trained ResNet-12 embedding module, we notice that the performance of ProtoNet is superior to that of RelationNet, which is consistent with the conclusion in \cite{closer_iclr19}. Note that there are no extra parameters in the metric learning module when negative Euclidean distance is used like ProtoNet.
For the ResNet-12 embedding module, its structure and parameter are designed and pre-trained well for static appearance feature extraction. The extracted feature map has a large number of channels and a low resolution, which increases the difficulty of relation modeling. However, the proposed method still helps a lot for better learning ability.
The consistent improvement by combining with our method demonstrates the effectiveness of the proposed method.

\begin{table}[!t]
\caption{Few-shot learning accuracies on miniImageNet.
\# denotes the 30-way for 1-shot and 20-way for 5-shot training method in \cite{Prototypical_NIPS17}. $\star$ indicate the implementation result reproduced by us.
}
\label{table:miniImageNet}
\centering
\begin{tabular}{l|c|c}
\hline
\hline
Model & 5-way 1-shot & 5-way 5-shot  \\
\hline
MatchNet \cite{Matching_Networks_NIPS16}  & 46.6\% &  60.0\%  \\                    
Meta-Learner LSTM \cite{ravi2016optimization} & 43.44+-0.77\% & 60.60+-0.71\% \\
MetaNet \cite{munkhdalai2017meta} & 49.21+-0.96\% & - \\
MAML \cite{MAML_ICML17} & 48.70+-1.84\% & 63.11+-0.92\% \\
ProtoNet \cite{Prototypical_NIPS17}  & 46.14+-0.77\% & 65.77+-0.70\% \\
ProtoNet\# \cite{Prototypical_NIPS17} & \textit{49.42}+-0.78\% & \textit{68.20}+-0.66\% \\
RelationNet \cite{Compare_CVPR18}  &  50.44+-0.82\%  &  65.32+-0.70\%  \\
RelationNet$\star$ \cite{Compare_CVPR18}  &  49.80+-0.82\%  & 64.71+-0.69\%   \\
GNN \cite{GNN_iclr18} & 50.33+-0.36\% & 66.41+-0.63\% \\
CovaMNet \cite{li2019distribution} & 51.19+-0.76\% & 67.65+-0.63\% \\
R2D2 \cite{R2D2_iclr19}  & 51.8+-0.2\% & 68.4+-0.2\% \\
Sampler-FC  \cite{patch_sampler_CVPR19}  & 47.18+-0.83\%  &  66.41+-0.67\%  \\
Sampler-CS  \cite{patch_sampler_CVPR19}  & 51.03+-0.78\%  &  67.96+-0.71\%  \\
Localization \cite{CVPR2019_localization}  &  \textit{49.64+-0.31\%}  &  \textit{69.45+-0.28\%}  \\
SoSN \cite{WACV2019_powernorm}  &  52.96+-0.83\%  &  68.63+-0.68\%  \\
SoSN+Permutations \cite{WACV2019_powernorm}  &  \textit{54.72+-0.89\%}  &  \textit{68.67+-0.67\%}  \\
DN4 \cite{Local_Descriptor_CVPR19}  &  51.24+-0.74\%  &  71.02+-0.64\%  \\
SAML \cite{SAML_ICCV19} & 52.22\% & 66.49\% \\
SAML \cite{SAML_ICCV19} (224$\times$224 input) & 56.68+-0.40\% & 71.34+-0.41\% \\
FEAT \cite{ye2020few} (e=10000) & 55.15+-0.20\% & \textbf{71.61}+-0.16\%  \\
\hline
Ours  &  53.33+-0.88\%  &  68.92+-0.69\% \\    
Ours (224$\times$224 input)  &  \textbf{56.71}+-0.90\%  &  71.50+-0.71\% \\    
\hline
TADAM \cite{adaptive_metric_NIPS18} (ResNet-12)  &  58.5\%  &  76.7\%  \\
MTL \cite{meta_transfer_CVPR19} (ResNet-12)  &  61.2+-1.8\%  &  75.5+-0.8\%  \\
LEO \cite{RusuRSVPOH19} (WRN-28-10, e=50000)  &  61.76+-0.08\%  &  77.59+-0.12\%  \\
CTM \cite{Task_Relevant_CVPR19} (ResNet-18)  &  62.05+-3.84\%  &  78.63+-4.34\%  \\
MetaOptNet \cite{lee2019meta} (ResNet-12)  &  62.64+-0.61\%  &  78.63+-0.46\%  \\
FEAT \cite{ye2020few} (ResNet-12, e=10000) & \textbf{66.78}+-0.20\% & \textbf{82.05}+-0.14\%  \\
\hline
Ours (ResNet-12, e=10000) & 63.47+-0.20\% & 81.27+-0.15\% \\
\hline
\hline
\end{tabular}
\end{table}

\begin{table}[!t]
\caption{Few-shot learning accuracies on tieredImageNet.}
\label{table:tieredImageNet}
\centering
\begin{tabular}{l|c|c}
\hline
\hline
Model & 5-way 1-shot & 5-way 5-shot  \\
\hline
Masked Soft k-Means \cite{Task_Relevant_CVPR19}  &  52.39+-0.44\%  &  69.88+-0.20\%  \\
SoSN \cite{WACV2019_powernorm}  &  58.62+-0.92\%  &  75.19+-0.79\%  \\
CTM \cite{Task_Relevant_CVPR19} (ResNet-18)  &  64.78+-3.67\%  &  81.05+-3.66\%  \\
MetaOptNet \cite{lee2019meta} (ResNet-12)  &  65.99+-0.72\%  &  81.56+-0.53\%  \\
LEO \cite{RusuRSVPOH19} (WRN-28-10, e=50000)  &  66.33+-0.05\%  &  81.44+-0.09\%  \\
FEAT \cite{ye2020few} (ResNet-12, e=10000) & \textbf{70.80}+-0.23\% & 84.79+-0.16\%  \\
\hline
Ours (ResNet-12, e=10000) & 68.58+-0.23\% & \textbf{84.92}+-0.16\%  \\
\hline
\hline
\end{tabular}
\end{table}

\begin{table*}[!t]
\caption{Few-shot learning accuracies on Omniglot.
\# denotes the 60-way training method in \cite{Prototypical_NIPS17}, which is different from the setting of other methods. $\star$ indicate the implementation result reproduced by us.
}
\label{table:omniglot}
\centering
\begin{tabular}{l|c|c|c|c}
\hline
\hline
Model & 5-way 1-shot & 5-way 5-shot & 20-way 1-shot & 20-way 5-shot \\
\hline
MANN \cite{MANN_ICML16} & 82.8\% & 94.9\% & - & \\
Conv-Siamese Net \cite{siamese_ICMLW15} & 97.3\% & 98.4\% & 88.1\% & 97.0\% \\
MatchNet \cite{Matching_Networks_NIPS16}  & 98.1\% &  98.9\%  &  93.8\%  &  98.5\%  \\
Siamese with Memory \cite{DBLP:conf/iclr/KaiserNRB17} & 98.4\% & 99.6\% & 95.0\% & 98.6\% \\
Neural Statistician \cite{DBLP:conf/iclr/EdwardsS17} & 98.1\% & 99.5\% & 93.2\% & 98.1\% \\
MetaNet \cite{munkhdalai2017meta} & 99.0\% & - & 97.0\% & - \\
ProtoNet \cite{Prototypical_NIPS17}  & 97.4\% & 99.3\% & 95.4\% & 98.7\% \\
ProtoNet\# \cite{Prototypical_NIPS17} & 98.8\% & 99.7\% & 96.0\% & 98.9\% \\
MAML \cite{MAML_ICML17} & 98.7+-0.4\% & \textbf{99.9}+-0.1\% & 95.8+-0.3\% & 98.9+-0.2\%  \\
RelationNet \cite{Compare_CVPR18}  &  99.6+-0.2\%  &  99.8+-0.1\%  &  97.6+-0.2\%  &  99.1+-0.1\%  \\
RelationNet$\star$ \cite{Compare_CVPR18}  &  99.51+-0.20\%  &  99.75+-0.07\%  &  97.17+-0.23\%  &  99.01+-0.08\%  \\
Sampler-FC  \cite{patch_sampler_CVPR19}  & 97.43+-0.28\%  &  99.51+-0.07\%  & - & - \\
Sampler-CS  \cite{patch_sampler_CVPR19}  & 97.56+-0.31\%  &  99.65+-0.06\%  & - & - \\
Two-Stage \cite{das2019two} & 99.2+-0.3\% & 99.5+-0.2\% & 97.2+-0.3 & 98.9+-0.3 \\
\hline
Ours  &  \textbf{99.70}+-0.20\%  &  99.82+-0.09\%  & \textbf{97.49}+-0.22\%  & \textbf{99.15}+-0.08\%  \\
\hline
\hline
\end{tabular}
\end{table*}

\begin{table}[!t]
\caption{Few-shot learning accuracies on CUB. * denotes that the result is cited from \cite{closer_iclr19} which is trained with data augmentation. $\star$ indicate the implementation result reproduced by us.
}
\label{table:CUB}
\centering
\begin{tabular}{l|c|c}
\hline
\hline
Model & 5-way 1-shot & 5-way 5-shot  \\
\hline
MatchNet* \cite{Matching_Networks_NIPS16}  & 60.52+-0.88\% &  75.29+-0.75\%  \\       
MAML* \cite{MAML_ICML17} & 54.73+-0.97\% & 75.75+-0.76\% \\
ProtoNet* \cite{Prototypical_NIPS17} & 50.46+-0.88\% & 76.39+-0.64\% \\
mAP-SSVM \cite{triantafillou2017few} & 59.0\% & - \\
mAP-DLM  \cite{triantafillou2017few} & 59.1\% & - \\
RelationNet* \cite{Compare_CVPR18}  &  62.34+-0.94\%  & 77.84+-0.68\%   \\
RelationNet$\star$ \cite{Compare_CVPR18}  &  61.84+-1.05\%  & 77.31+-0.72\%   \\
Baseline \cite{closer_iclr19} & 47.12+-0.74\% & 64.16+-0.71\% \\
Baseline++ \cite{closer_iclr19} & 60.53+-0.83\% & 79.34+-0.61\% \\
SAML \cite{SAML_ICCV19} & 69.33+-0.22\% & 81.56+-0.15\% \\  
FEAT \cite{ye2020few} (e=10000) & 68.87+-0.22\% & \textbf{82.90}+-0.15\% \\
\hline
Ours (e=10000) &  \textbf{69.57}+-0.25\% & 81.73+-0.16\% \\
\hline
\hline
\end{tabular}
\end{table}

\begin{table}[!t]
\caption{Few-shot learning accuracies on Flower-102 and Food-101 datasets with 5-way 1-shot and 5-shot settings.}
\label{table:Flower_Food}
\centering
\begin{tabular}{l|c c|c c}
\hline
\hline
\multirow{2}*{Model} & \multicolumn{2}{c|}{Flower-102} & \multicolumn{2}{c}{Food-101} \\
~ & 1-shot & 5-shot & 1-shot & 5-shot  \\
\hline
ProtoNet \cite{Prototypical_NIPS17} & 62.81\% & 82.11\% & 36.71\% & 53.43\% \\
RelationNet \cite{Compare_CVPR18}  &  68.52\%  & 81.11\% & 36.89\% & 49.07\%   \\
SoSN \cite{WACV2019_powernorm} & 76.27\% & 88.55\% & 43.12\% & 58.13\% \\
MsSoSN (3 scales) \cite{zhang2020few} & 81.65\% & 92.10\% & 50.87\% & 66.32\% \\
MsSoSN+SS+SD+DD \cite{zhang2020few} & 84.69\% & 94.21\% & 53.86\% & 68.67\% \\
\hline
Ours (ResNet-12)  &  \textbf{84.92}\%  &  \textbf{94.52}\%  &  \textbf{56.21}\%  &  \textbf{73.94}\%  \\
\hline
\hline
\end{tabular}
\end{table}

\subsection{Comparison with State-of-the-Art}

We compare our method with the state-of-the-art approaches on miniImageNet, tieredImageNet, Omniglot, CUB, Flower-102 and Food-101 datasets with several learning settings.
Although our method can be combined with multiple state-of-the-art models, without loss of generality, we use RelationNet based on shallow embedding module and ProtoNet based on deep embedding module for comparison.

\subsubsection{Results on miniImageNet and tieredImageNet datasets}
The results on the miniImageNet dataset are shown in Table \ref{table:miniImageNet}.
Among the methods based on 4-layer convolutional embedding modules, our proposed method outperforms most of the other methods significantly. The comparison with some representative methods are analyzed in detail as follows.

Localization \cite{CVPR2019_localization} and SoSN \cite{WACV2019_powernorm} also used second-order statistics for few-shot learning. However, the Localization model \cite{CVPR2019_localization} localized objects using bounding box annotations before classification and computed the cross-covariance between the predicted foreground and background maps, which helped a lot in performance improvement.
SoSN made a thorough analysis in second-order statistics and power normalization functions. The derived sigmoid-like power normalizing function can improve the performance of 5-way 1-shot and 5-shot learning on miniImageNet from 50.88\% and 66.71\% to 52.96\% and 68.63\%, which demonstrates its effectiveness and provides insight for the society. They also proposed to permute the second-order matrices to capture various correlations with multiple similarity networks, which can be seen as data augmentation.
We propose to learn and compare appearance and relation features separately and complementarily to solve content misalignment for few-shot learning. We use the first-order static appearance information, the second-order correlation information, and the local-global mutual information to constrain the model to learn consistent and intrinsic features. The learning procedure can be regarded as optimization with multiple constraints. In our relation stream, we enforce the features of the same category to share a common correlation matrix. Both the relation among spatial local features and the relation among channel style features are taken into account, while other works only considered the first one. It should be noted that there is only 1 fully-connected layer to estimate the matching score based on the difference of the correlation matrices. And we do not use any version of extra annotations, power normalizations or permutation based augmentations in our method.

DN4 \cite{Local_Descriptor_CVPR19} and SAML \cite{SAML_ICCV19} are based on local descriptors for similarity computation.
DN4 conducted online k-nearest neighbor search over the deep local descriptors.
Under the 1-shot setting, our proposed method achieves 53.33\% accuracy, which is more than 2\% higher than that of DN4. Under the 5-shot setting, DN4 performs better. The performance of DN4 depends heavily on the number of samples and the value of k. As the number of shots increases, the number of local descriptors to be compared increases, hence its performance gets better and better. However, the computational complexity also increases. Different from DN4 searching from the entire support set, we assemble the shots to serve as prototypes for each class like ProtoNet, which reduces complexity while lost some information.
SAML calculated the similarity of each local region pairs and used attention to aggregate the pairs' similarities for image comparison. Besides the $84\times84$ input, they also reported the recognition accuracy with $224\times224$ input. Our method outperforms SAML in all the settings. Simply using $224\times224$ input instead of $84\times84$ input, the performance of our method can yields up to 3\% improvement, yielding the highest accuracies for both 1-shot and 5-shot learning except FEAT \cite{ye2020few}.

Instead of using the same embedding function for all tasks, FEAT \cite{ye2020few} proposed to adapt the embedding to each target few-shot learning task with a Transformer based set-to-set function, which improves the performance significantly. They also used a more trustworthy evaluation setting with 10,000 sampled tasks instead of 600 sampled tasks. We follow to use this evaluation in the experiments based on the deep embedding module. As shown in the lower part of Table \ref{table:miniImageNet}, among the methods based on deep embedding modules, our proposed method outperforms all the other methods except FEAT.
The result on the tieredImageNet dataset is consistent with that on the miniImageNet dataset as shown in Table \ref{table:tieredImageNet}.
Our method even surpasses FEAT under 5-way 5-shot setting on the tieredImageNet.
CTM \cite{Task_Relevant_CVPR19} and MetOptNet \cite{lee2019meta} also shared the same spirit of learning task-specific features, whose performances are inferior to mine.
As analyzed above, the deep embedding module is designed and pre-trained well for appearance feature extraction, hence the accuracy is boosted drastically compared to shallow models. However, the extracted feature map with the deep embedding module has a large number of channels and a low resolution, which increases the difficulty of relation modeling. In the future we will explore more about the structure of relation modeling and combination with embedding adaptation methods to further promote the performance.



\subsubsection{Results on Omniglot and CUB datasets}
We compare our method with other representative methods on the Omniglot dataset and CUB dataset, whose results are listed in Table \ref{table:omniglot} and Table \ref{table:CUB} respectively.
It should be noted that, on the CUB dataset, the performance of our method is on par with FEAT under 5-way 5-shot setting, and even better than FEAT by 0.7\% under 1-shot setting. The experimental results demonstrate the effectiveness of our method, especially for fine-grained dataset.
The proposed method not only pays more attention to the relation information inside images, which is insensitive to the absolute value, but it can also encourage the network to learn locally-consistent essential features shared among images from the same class. As a result, it allows for high representation quality and a generic system robust to content misalignment and noise.

\subsubsection{Results on Flower-102 and Food-101 datasets}
To further support the conclusion, we evaluate our method on another two fine-grained datasets, \textit{i.e.}, Flower-102 and Food-101. The results are listed in Table \ref{table:Flower_Food}.
MsSoSN \cite{zhang2020few} is a multi-scale version of SoSN, which not only used $84\times84$ input, but also used $64\times64$ and $256\times256$ inputs. MsSoSN+SS+SD+DD represent adding a scale selector, a scale discriminator, and a discrepancy discriminator to MsSoSN in addition. Instead of using multi-scale inputs, we report the performance of our method based on the ResNet-12 embedding module and $84\times84$ input. Our method achieves a new state-of-the-art performance on these datasets.


\section{Conclusion}

In order to address content misalignment for few-shot learning, we propose a novel semantic alignment model with multiple streams to compare relations as well as for better representation and metric learning ability.
We introduce a relation stream to align and compare correlation relations among the elements inside an image.
To take both the relationship of positional features and the relationship of style features into account, we apply constraint to the correlation matrix across spatial positions as well as the correlation matrix across channels.
Besides that, the local-to-global consistency relation is optimized with a mutual information stream. This stream plays an important role in the quality of the representation learning. Locally-consistent and intra-class shared features are encouraged.
The two proposed streams not only perform well by themselves but also can be fused with the existing appearance comparison methods mutual reinforcingly with learnable weights, demonstrating the effectiveness of our proposed method and indicating that semantic relation robust to misalignment is an important factor in few-shot learning.



%
%
%

%

\bibliographystyle{IEEEtran}
\bibliography{IEEEabrv,mybibfile}

\end{document}